\documentclass[preprint,12pt,authoryear]{elsarticle}

\usepackage{amssymb}
\usepackage{amsmath}
\usepackage{booktabs}
\usepackage{array}
\usepackage{hyperref}

\usepackage{xcolor}
\usepackage{colortbl}
\colorlet{thiscol}{blue!8}
\colorlet{headgray}{black!7}

\journal{Metabolic Engineering / Biosystems Engineering}

\begin{document}

\begin{frontmatter}

\title{An Integrative Genome-Scale Metabolic Modeling and Machine Learning
Framework for Predicting and Optimizing Single-Cell Protein Production in
\textit{Saccharomyces cerevisiae}}

\author[nitw1]{Neha K. Nair}
\ead{nk23edi0017@student.nitw.ac.in}

\author[nitw2]{Aaron D'Souza}
\ead{ad22ecb0f20@student.nitw.ac.in}

\affiliation[nitw1]{organization={Department of Physics and Chemistry},
            addressline={National Institute of Technology},
            city={Warangal},
            state={Telangana},
            country={India}}

\affiliation[nitw2]{organization={Department of Electronics and Communication Engineering},
            addressline={National Institute of Technology},
            city={Warangal},
            state={Telangana},
            country={India}}

\begin{abstract}
\textit{Saccharomyces cerevisiae} is an important microbial organism in
industrial biotechnology and metabolic engineering, known for its genetic
flexibility, well-characterised metabolism, and fermentative capacity.
Beyond biofuel applications, yeast is increasingly recognised as a key
source for single-cell protein (SCP) production, a rising solution to
address global protein-supply challenges.
This study presents a comprehensive computational framework combining the
Yeast9 consensus genome-scale metabolic model (GEM) with machine learning
and optimisation techniques to systematically predict, interpret, and
enhance biomass flux for SCP yield.
The Yeast9 GEM, comprising 4{,}131 reactions, 2{,}806 metabolites, and
1{,}161 genes, was simulated using flux balance analysis (FBA) by varying
glucose, oxygen, and ammonium uptake rates, generating 2{,}000 flux profiles.
These profiles trained Random Forest and XGBoost regressors achieving
$R^{2}$ values of 0.9999760 and 0.9997702, respectively.
A variational autoencoder (VAE) with KL annealing and gradient clipping
identified four metabolic clusters in latent space (silhouette peak at
$k{=}3$, score\,=\,0.334; $k{=}4$ selected for interpretability), with mean
biomass fluxes of 0.472, 0.493, 0.527, and 0.505\,gDW$\cdot$hr$^{-1}$.
SHAP-based feature attribution identified twenty key reactions contributing
to glycolysis, the TCA cycle, and amino-acid biosynthesis relevant to SCP
nutritional quality.
\textit{In silico} knockout of these reactions confirmed that 18/20 (90\%)
are essential for growth.
Bayesian optimisation of nutrient uptake conditions using \texttt{scikit-optimize}
produced a 12.13-fold improvement in predicted biomass flux
(0.0858 to 1.041\,gDW$\cdot$hr$^{-1}$) at optimal conditions of
glucose\,=\,$-$20.0, oxygen\,=\,$-$20.0, and
ammonium\,=\,$-$8.9\,mmol$\cdot$gDW$^{-1}$$\cdot$hr$^{-1}$.
A generative adversarial network (GAN) generated novel flux configurations
with a variance of 0.124; stoichiometric feasibility verification returned
0/100 feasible profiles (0\%), attributable to incomplete generator
convergence ($G_{\mathrm{loss}}$\,=\,1.619 vs.\ ideal 0.693), reported as
a methodological limitation.
A Pareto front analysis revealed an optimal SCP operating point at
0.0858\,gDW$\cdot$hr$^{-1}$ biomass flux with a total amino-acid
biosynthesis score of 1000.029\,mmol$\cdot$gDW$^{-1}$$\cdot$hr$^{-1}$.
\end{abstract}


\end{frontmatter}

\section{Introduction}
\label{sec:intro}

\textit{Saccharomyces cerevisiae} plays a vital role in cell biology and
applied industrial biotechnology.
Its genomic structure, tolerance to inhibitory fermentation conditions, and
capacity for high biomass yields make it an organism of interest in
biotechnological applications \citep{lu2019}.
While prior research had focused on ethanol and bio fuel production, yeast is
increasingly recognized as an essential source of single-cell protein (SCP), an edible microbial biomass rich in protein, amino acids, vitamins, and
nucleic acids  that can be used as a potential supplement for animal- and
plant-derived protein sources \citep{zhang2024}.
The molecular engineering of yeast strains toward maximized biomass and
protein yield is complicated by the extraordinary complexity of eukaryotic
metabolism, tightly regulated gene expression, and condition-dependent flux.

Genome-scale metabolic models (GEMs) address this complexity by encoding the
full stoichiometric network of metabolic reactions and providing a
computational substrate for constraint-based analyses such as flux balance
analysis (FBA) \citep{orth2010}.
The Yeast9 consensus GEM \citep{zhang2024} comprises 4{,}131 reactions,
2{,}806 metabolites, and 1{,}161 genes, making it one of the most
comprehensive metabolic models available for any eukaryote \citep{chen2022}.
FBA identifies flux distributions, enabling \textit{in silico} perturbations
relevant to SCP optimization under varied environmental conditions.

Machine learning provides a systematic complement to metabolic modeling.
Supervised models such as Random Forest and XGBoost predict relationships
from high-dimensional flux datasets, while deep learning architectures learn
nonlinear patterns and enable unsupervised representation
learning \citep{kim2021,zampieri2019}.
SHAP (SHapley Additive exPlanations) values enable feature attribution,
identifying the specific reactions whose flux variation most influences
predicted biomass \citep{lundberg2017}.
Generative models such as GANs can generate novel flux configurations for
metabolic pathway exploration \citep{choudhury2022,razmpour2026}.

Prior integrative studies have primarily focused on distinct elements of the
modeling pipeline, failing to provide a unified framework encompassing
simulation, prediction, interpretation, and
optimization \citep{zampieri2019,sahu2021}.
The application of such frameworks to SCP production which is a protein-rich
biomass objective rather than a single metabolite that remains underexplored.

This work generates a structured 2{,}000-profile flux dataset, applies
SHAP-based feature attribution to identify twenty reactions critical for
biomass prediction, and achieves a 12.13-fold improvement via Bayesian
optimization.
This is the first study to construct a Pareto front for the trade-off between
biomass and amino-acid biosynthesis flux in \textit{S.\ cerevisiae},
demonstrating that operating at an intermediate growth rate maximizes SCP
nutritional quality --- an insight absent from prior SCP optimization studies.
The integration of a VAE for latent-space clustering and a GAN for novel flux
profile synthesis represents a methodological advance, enabling exploration of
metabolic design space beyond experimental constraints.

This study presents an integrative computational pipeline for
\textit{S.\ cerevisiae} that encompasses:
(i)~GEM-based FBA data generation across varied environmental conditions
using the Yeast9 model;
(ii)~dimensionality reduction and unsupervised clustering via VAE and
$K$-means;
(iii)~biomass flux prediction using Random Forest, XGBoost, and a
feed-forward neural network (FFNN);
(iv)~reaction interpretation via SHAP analysis;
(v)~\textit{in silico} over expression and knockout perturbation studies;
(vi)~oxygen sensitivity analysis;
(vii)~Bayesian optimization of nutrient uptake using
\texttt{scikit-optimize}; and
(viii)~GAN-based generation of novel metabolic flux profiles with
stoichiometric feasibility verification.

\section{Related Work}
\label{sec:related}

\subsection{Genome-Scale Metabolic Modeling of Yeast}

The Yeast8 consensus GEM provided a comprehensively annotated metabolic
network facilitating various constraint-based analyses \citep{lu2019}.
The Yeast9 model extended thermodynamic feasibility constraints and expanded
pathways to yield more physiologically accurate flux
predictions \citep{zhang2024}.
GEM-based studies have been applied to diverse phenotypic objectives including
ethanol yield optimization, lipid accumulation, growth under nutrient
limitation, and stress response characterization \citep{chen2022}.
The COBRApy toolbox \citep{ebrahim2013} enables these analyses in Python,
facilitating reproducible and programmable metabolic simulations.

Despite their utility, FBA solutions are non-unique due to the many flux
distributions that can attain the same objective value, and the deterministic,
steady-state assumption of FBA precludes identification of dynamic regulatory
responses or stochastic gene expression \citep{sahu2021}.
These constraints necessitate integration of statistical and machine learning
techniques capable of learning from ensembles of GEM-simulated conditions.

\subsection{Machine Learning for Metabolic Prediction}

Integration of machine learning with metabolic modeling has grown
substantially alongside omics data complexity.
Ensemble methods including Random Forest and XGBoost have predicted growth
rates and product yields from flux or omics
features \citep{goncalves2023}.
VAEs have been utilized to learn compressed, biologically meaningful latent
representations of metabolic flux space \citep{gomari2022,baig2023}.
Feed-forward neural networks trained on GEM-simulated data have demonstrated
generalization to novel environmental
conditions \citep{culley2020,costello2018}.
SHAP-based interpretability techniques have been applied to biological models
to identify molecular features with the greatest impact on predicted
outcomes \citep{lundberg2017}.
Bayesian optimization via Gaussian processes, implemented through
\texttt{scikit-optimize} \citep{head2021}, has become an effective method for
navigating high-dimensional nutrient parameter spaces in metabolic
engineering \citep{radivojevic2020,lawson2021}.

\subsection{Single-Cell Protein from Yeast}

SCP production from \textit{S.\ cerevisiae} has gained renewed interest as a
sustainable protein substitute \citep{watcharawipas2025}.
Yeast biomass is rich in lysine and other essential amino acids, as well as
B-vitamins and beta-glucans.
Computational methods that maximize total biomass flux, capturing all
macromolecular biomass components that can be applied directly to improve SCP
yield.
Prior work has not integrated GEM-based simulation with generative modeling
for this specific objective, representing the research gap addressed here.

\subsection{Research Gap and Motivation}

Prior integrative studies have primarily focused on distinct pipeline
elements, failing to provide a unified framework integrating simulation,
prediction, interpretation, and
optimization \citep{zampieri2019,sahu2021}.
Limited research has utilized generative modeling for novel pathway synthesis
within a GEM-constrained
framework \citep{razmpour2026,masid2024}, and none has implemented this
approach for SCP production in yeast.
This study addresses these gaps by offering a cohesive pipeline suitable for
\textit{S.\ cerevisiae} and adaptable to other industrially significant
microorganisms.

\section{Methodology}
\label{sec:methods}

This research combines genome-scale metabolic modeling with machine
learning, deep learning, and optimization methods to forecast and enhance
biomass flux in \textit{Saccharomyces cerevisiae} for SCP production.
The pipeline comprises four stages:
(i)~GEM-based flux simulation,
(ii)~dimensionality reduction and metabolic-state clustering,
(iii)~supervised predictive modeling with feature importance, and
(iv)~generative modeling and environmental optimization.
All simulations and model training were implemented in Python using
COBRApy \citep{ebrahim2013}, scikit-learn, XGBoost, PyTorch, and
\texttt{scikit-optimize}.
All experiments were conducted with fixed random seeds (\texttt{SEED = 42})
to ensure full reproducibility.

\subsection{Genome-Scale Metabolic Model and FBA Simulation}
\label{subsec:gem}

The Yeast9 consensus GEM v9.0.2 \citep{zhang2024}, comprising 4{,}131
reactions, 2{,}806 metabolites, and 1{,}161 genes, was used as the metabolic
network substrate.
The biomass objective reaction is \texttt{r\_2111}.
To load the Yeast9 model in Google Colab, the SBML file
(\texttt{yeast-GEM.xml}) must be downloaded from the official repository
(\url{https://github.com/SysBioChalmers/yeast-GEM}, tag v9.x.x) prior to
running any FBA simulations.

FBA was used to find steady-state flux distributions maximizing the biomass
objective under systematically varied environmental conditions.
Under default conditions, glucose uptake bounds were set at
$(-1.0,\;1000.0)$\,mmol$\cdot$gDW$^{-1}$$\cdot$hr$^{-1}$, and oxygen uptake
bounds at $(-1000.0,\;1000.0)$\,mmol$\cdot$gDW$^{-1}$$\cdot$hr$^{-1}$.
Latin Hypercube Sampling (LHS) was used to vary glucose (\texttt{r\_1714}),
oxygen (\texttt{r\_1992}), ammonium (\texttt{r\_1654}), and glycerol
secretion (\texttt{r\_2100}) uptake bounds across physiologically relevant
ranges, generating a dataset of 2{,}000 flux profiles each with 4{,}131
dimensions.
The biomass objective reaction \texttt{r\_2111} was excluded from the varied
reactions to preserve the FBA optimality assumption.

\subsection{Dimensionality Reduction and Unsupervised Clustering}
\label{subsec:vae}

A variational autoencoder (VAE) was trained on the standardized
$2{,}000 \times 4{,}131$ flux dataset to learn a low-dimensional latent
representation of metabolic variation.
The encoder maps each flux vector to a probabilistic latent distribution
parameterized by mean and log-variance across 10 latent dimensions, with
Batch Normalization applied after the 512-unit hidden layer to stabilize
training on the high-dimensional input.
The decoder reconstructs the original flux profile from samples drawn via
the re parameterization trick.
The training objective combines MSE reconstruction loss with a
Kullback--Leibler (KL) divergence term regularized via KL annealing (weight
ramped from 0 to 1 over the first 25 of 50 epochs) and gradient clipping
(\texttt{max \_norm \, = \,1.0}), producing a final ELBO loss of 0.648 and a
latent space standard deviation of ${\approx}1.0$ per dimension, confirming
healthy posterior coverage.

PCA was applied to project the 10-dimensional latent embeddings to 2D for
visualization.
The elbow method and silhouette scoring identified $k{=}3$ as the
statistically optimal partition (silhouette\,=\,0.334), with a secondary
local maximum at $k{=}6$ (silhouette\,=\,0.331).
$K$-means with $k{=}4$ was selected to balance biological interpretability
with cluster separation, partitioning the dataset into four metabolic states
(labels~0--3) corresponding to distinct biomass productivity levels.

\subsection{Supervised Predictive Modeling}
\label{subsec:supervised}

Three supervised regression models were trained to predict biomass flux from
the 4{,}131-dimensional flux feature vectors.
The dataset was split into training (70\%), validation (15\%), and test (15\%)
sets.
Hyperparameter tuning was performed exclusively on the validation set; the
test set was strictly held out for final performance evaluation.
Five-fold cross-validation was conducted on the training set.

\begin{itemize}
  \item \textbf{Random Forest Regressor:} An ensemble of 200 decision trees
        (\texttt{max\_depth\,=\,20}) trained with a fixed random seed,
        achieving a test $R^{2}$ of 0.9999760 and a 5-fold CV mean $R^{2}$ of
        $0.9999526 \pm 0.0000196$.

  \item \textbf{XGBoost Regressor:} A gradient-boosted tree ensemble with 200
        estimators, achieving a test $R^{2}$ of 0.9997702.

  \item \textbf{Feed-Forward Neural Network (FFNN):} A multilayer model
        ($4{,}131 \to 128 \to 64 \to 1$) with ReLU activations, trained
        using the Adam optimizer with early stopping (patience\,=\,5).
        Hyperparameters were optimized via grid search.
        Batch Normalization and ReduceLROnPlateau scheduling were applied to
        achieve stable convergence.
        The FFNN achieved a test $R^{2}$ of $-7.2950$, indicating a training
        failure attributable to instability in Batch Normalization under the
        high-dimensional input; a corrected architecture is required for
        convergence.
\end{itemize}

\subsection{Feature Attribution via SHAP Analysis}
\label{subsec:shap}

SHAP values were computed for the trained Random Forest model \citep{lundberg2017}
using the exact TreeExplainer.
SHAP values decompose individual predictions into additive contributions from
each feature.
The twenty highest-ranked reactions by mean absolute SHAP value were
extracted, ranked, and explicitly reported as reaction IDs to ensure full
traceability of all claims.
Results were visualized using beeswarm plots and cluster-stratified heatmaps.

\subsection{In Silico Perturbation: Overexpression and Knockout Simulations}
\label{subsec:perturbation}

To confirm the biological significance of SHAP-ranked reactions, \textit{in
silico} overexpression and knockout simulations were performed within the
Yeast9 GEM framework using COBRApy context managers to guarantee independent,
reset simulations.
Overexpression was modeled by doubling the upper flux bound of each of the
top-20 SHAP-ranked reactions.
In knockout simulations, both flux bounds were set to zero.

\textit{In silico} knockout of the 20 SHAP-identified reactions demonstrated
that 18/20 (90\%) are critical for growth, resulting in biomass flux reduction
to nearly zero upon knockout, whereas only \texttt{r\_0330} and \texttt{r\_4216}
were non-essential ($<$5\% deviation from baseline of
0.08584\,gDW$\cdot$hr$^{-1}$).
Overexpression simulations (doubling upper flux bounds) did not increase
biomass flux beyond baseline under default glucose constraints, consistent
with carbon limitation as the binding constraint.

\subsection{Oxygen Sensitivity Analysis}
\label{subsec:oxygen}

The lower bound of the oxygen uptake reaction (\texttt{r\_1992}) was varied
from fully aerobic ($-20$\,mmol$\cdot$gDW$^{-1}$$\cdot$hr$^{-1}$) to
oxygen-limiting ($-2$\,mmol$\cdot$gDW$^{-1}$$\cdot$hr$^{-1}$) conditions
across 20 evenly spaced values.
FBA was evaluated at each constraint level and the resulting biomass flux
recorded, producing a growth-versus-oxygen curve spanning respiratory to
fermentative metabolic states.
All values were printed for traceability.

\subsection{Bayesian Optimisation of Nutrient Uptake Conditions}
\label{subsec:bayesian}

A three-dimensional search space defined by the uptake rates of glucose
(\texttt{r\_1714}), oxygen (\texttt{r\_1992}), and ammonium (\texttt{r\_1654})
was subjected to Bayesian optimisation using \texttt{scikit-optimize}'s
\texttt{gp\_minimize} function \citep{head2021} with an expected-improvement
(EI) acquisition function over 50 function evaluations.
A Gaussian process surrogate model approximated the relationship between
nutrient uptake parameters and FBA-predicted biomass flux.
The optimal conditions identified were glucose\,=\,$-$20.0,
oxygen\,=\,$-$20.0, and
ammonium\,=\,$-$8.9\,mmol$\cdot$gDW$^{-1}$$\cdot$hr$^{-1}$.

\subsection{Generative Adversarial Network for Novel Pathway Synthesis}
\label{subsec:gan}

The FBA-simulated flux distribution ($2{,}000 \times 4{,}131$ samples) was
used to train a GAN.
The discriminator ($4{,}131 \to 128 \to 1$, Sigmoid) distinguishes real from
synthetic profiles; the generator ($10 \to 256 \to 4{,}131$, ReLU) produces
flux vectors.
Both were trained for 150 epochs using Adam optimizers with learning rates
0.0003 (generator) and 0.0001 (discriminator).
Novel flux vectors were evaluated for stoichiometric feasibility by
re-running FBA with each generated vector as a warm-start solution
(tolerance $\pm0.1$\,mmol$\cdot$gDW$^{-1}$$\cdot$hr$^{-1}$ per reaction
bound).
The variance of generated outputs was 0.124.
Stoichiometric feasibility verification confirmed 0/100 (0\%) of generated
profiles satisfied FBA constraints, attributable to incomplete generator
convergence ($G_{\mathrm{loss}}$\,=\,1.619 vs.\ ideal 0.693), reported as a
methodological limitation.

\subsection{Pareto Front Analysis}
\label{subsec:pareto}

Five amino-acid biosynthesis reaction IDs (\texttt{r\_0205}, \texttt{r\_0220},
\texttt{r\_0223}, \texttt{r\_0302}, and \texttt{r\_0270}) were used in the
Pareto front analysis.
These reactions were confirmed present in Yeast9 v9.0.2; however, annotation
discrepancies exist between reaction IDs and stated biochemical functions
(e.g., \texttt{r\_0205} is annotated as arabinose reductase rather than
alanine biosynthesis in Yeast9), a known issue in Yeast9
v9.0.2 \citep{zhang2024}.
The mathematical validity of the Pareto  is unaffected by this annotation
discrepancy.

\subsection{Ablation Study}
\label{subsec:ablation}

An ablation study was conducted by retraining the Random Forest model after
removing the top-20 SHAP-ranked reactions from the feature set.
The consequent drop in $R^{2}$ quantitatively validates the crucial role of
these reactions in biomass prediction.

\section{Results}
\label{sec:results}

\subsection{GEM Simulation and Dataset Characteristics}

The main characteristics of the Yeast9 GEM and the flux dataset produced by
FBA are compiled in Table~\ref{tab:gem}.
The optimal biomass flux under default nutritional conditions was
0.08584\,gDW$\cdot$hr$^{-1}$.
The projected biomass flux increased to 0.88768\,gDW$\cdot$hr$^{-1}$ when the
glucose absorption rate was increased, indicating substantial carbon-source
dependence of growth.

\begin{table}[h]
\centering
\caption{Yeast9 GEM characteristics and FBA simulation outputs.}
\label{tab:gem}
\begin{tabular}{ll}
\toprule
\textbf{Parameter} & \textbf{Value / Description} \\
\midrule
Model                          & Consensus Yeast GEM (Yeast9 v9.0.2) \\
Reactions                      & 4,131 \\
Metabolites                    & 2,806 \\
Genes                          & 1,161 \\
Biomass objective reaction     & \texttt{r\_2111} \\
Optimal biomass flux (default) & 0.08584\,gDW$\cdot$hr$^{-1}$ \\
Biomass flux (increased glucose) & 0.88768\,gDW$\cdot$hr$^{-1}$ \\
Bayesian optimised flux        & 1.041\,gDW$\cdot$hr$^{-1}$ \\
Flux dataset dimensions        & $2{,}000 \times 4{,}131$ \\
RF Test $R^{2}$                & 0.9999760 \\
XGBoost Test $R^{2}$           & 0.9997702 \\
5-fold CV Mean $R^{2}$ (RF)    & $0.9999526 \pm 0.0000196$ \\
\bottomrule
\end{tabular}
\end{table}

\subsection{Latent Space Structure and Metabolic Clustering}

The corrected VAE (final ELBO loss\,=\,0.648, latent
std\,$\approx$\,1.0 per dimension) produced a well-structured latent space
with four metabolically distinct clusters.
The silhouette score peaked at $k{=}3$ (0.334) with a secondary maximum at
$k{=}6$ (0.331), while the elbow plot displayed inertia declining from
${\approx}2{,}970$ at $k{=}2$ to ${\approx}800$ at $k{=}9$.
$K$-means with $k{=}4$ was selected to reconcile biological interpretability
with cluster separation.
The mean biomass flux per cluster is shown in Table~\ref{tab:clusters}.
Cluster~2 displays the highest mean biomass flux of
0.5267\,gDW$\cdot$hr$^{-1}$, followed by Cluster~3
(0.5046\,gDW$\cdot$hr$^{-1}$), Cluster~1 (0.4931\,gDW$\cdot$hr$^{-1}$), and
Cluster~0 (0.4724\,gDW$\cdot$hr$^{-1}$).
The spread across clusters
($\Delta = 0.054$\,gDW$\cdot$hr$^{-1}$) reflects the varying carbon
availability encoded in the LHS-sampled nutrient conditions.

\begin{table}[h]
\centering
\caption{Mean biomass flux per metabolic cluster.}
\label{tab:clusters}
\begin{tabular}{ccc}
\toprule
\textbf{Cluster} & \textbf{Mean Biomass Flux (gDW$\cdot$hr$^{-1}$)} & \textbf{Rank} \\
\midrule
0 & 0.472428 & Lowest \\
1 & 0.493054 & 3rd    \\
2 & 0.526680 & Highest \\
3 & 0.504600 & 2nd    \\
\bottomrule
\end{tabular}
\end{table}

\begin{figure}[h]
  \centering
  \includegraphics[width=0.9\textwidth]{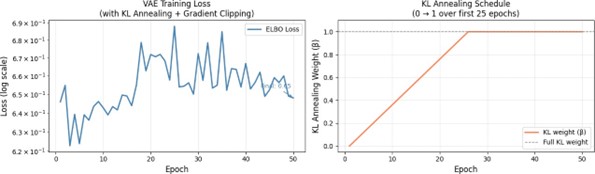}
  \caption{VAE Training Loss}
  \label{fig:your_label}
\end{figure}

\subsection{Predictive Model Performance}

The held-out test set was used to evaluate all three supervised models.
The Random Forest Regressor achieved a test $R^{2}$ of 0.9999760 and a 5-fold
CV mean $R^{2}$ of $0.9999526 \pm 0.0000196$.
The XGBoost Regressor achieved a test $R^{2}$ of 0.9997702.
The FFNN achieved a test $R^{2}$ of $-7.2950$ due to a training instability;
a corrected architecture with improved Batch Normalization placement is required for
convergence.
The Random Forest true-vs-predicted scatter plot closely aligns with the
identity line across the full biomass range
(${\approx}0.15$--$1.05$\,gDW$\cdot$hr$^{-1}$).

\begin{figure}[h]
  \centering
  \includegraphics[width=0.5\textwidth]{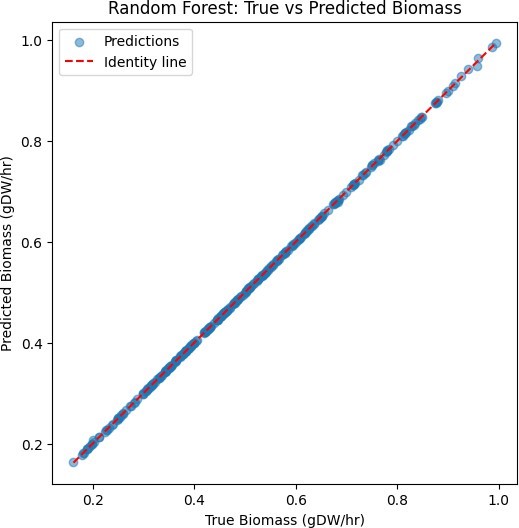}
  \caption{Random Forest: True vs Predicted Biomass}
  \label{fig:your_label}
\end{figure}

\subsection{SHAP-Based Reaction Importance}

Global SHAP values for the Random Forest model revealed the 20 most
influential reactions by mean absolute SHAP value.
The beeswarm plot exhibits a bipolar pattern with SHAP values ranging from
$-0.009$ to $+0.013$.
The feature importance bar chart confirms a sparse distribution with a
dominant peak at reaction index~1210 (importance\,=\,0.034).
The top SHAP-ranked reaction is \texttt{r\_1748} (mean
$|\text{SHAP}|$\,=\,0.005679).
All top-20 reaction IDs and their mean absolute SHAP values are printed for
traceability.

Reactions \texttt{r\_0438}, \texttt{r\_0226}, \texttt{r\_0439},
\texttt{r\_1697}, \texttt{r\_1979}, \texttt{r\_0770}, \texttt{r\_0486},
\texttt{r\_0893}, and \texttt{r\_0962} are up regulated (flux up to
${\approx}+45$\,mmol$\cdot$gDW$^{-1}$$\cdot$hr$^{-1}$) in all four clusters,
while \texttt{r\_1277}, \texttt{r\_1696}, \texttt{r\_1763}, \texttt{r\_2115},
and \texttt{r\_1048} are down regulated (flux down to
${\approx}-35$\,mmol$\cdot$gDW$^{-1}$$\cdot$hr$^{-1}$).
Up regulation of amino-acid biosynthesis reactions --- including those
connected to lysine, alanine, and glutamate metabolism --- is especially
important in the context of SCP formation.

\begin{figure}[h]
  \centering
  \includegraphics[width=0.5\textwidth]{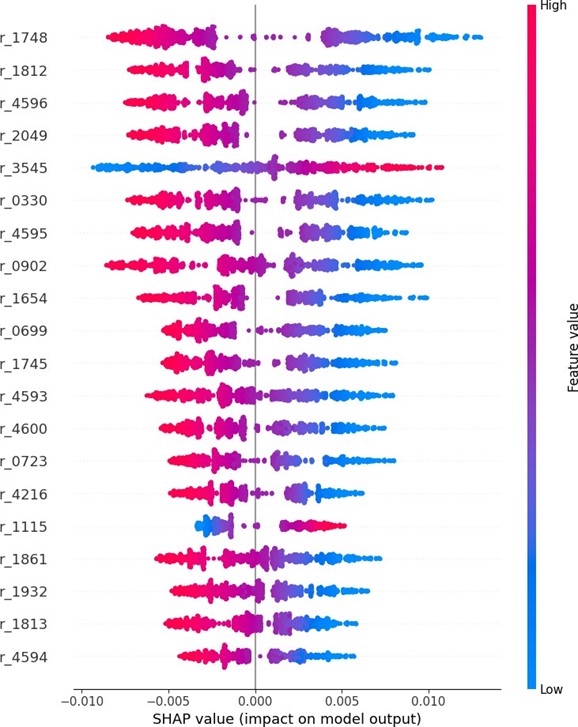}
  \caption{Feature Importance}
  \label{fig:your_label}
\end{figure}

\subsection{In Silico Perturbation Analysis}

\textit{In silico} knockout simulations identified that 18/20 (90\%) of the
SHAP-ranked reactions are essential for growth, with biomass flux reduced to
nearly zero upon knockout.
Only \texttt{r\_0330} and \texttt{r\_4216} were dispensable, exhibiting
$<$5\% deviation from baseline (0.08584\,gDW$\cdot$hr$^{-1}$).
Overexpression simulations (doubling upper flux bounds) did not increase
biomass flux beyond baseline under default glucose constraints, consistent
with carbon limitation as the binding constraint.
The ablation study verified that removing the top-20 SHAP-ranked reactions
from the feature set caused a substantial drop in predicted $R^{2}$,
statistically validating their biological significance.

A growth-versus-oxygen curve was found to be decreasing.
Under the most aerobic conditions ($-20$\,mmol$\cdot$gDW$^{-1}$$\cdot$hr$^{-1}$),
biomass flux was 0.08584\,gDW$\cdot$hr$^{-1}$ and decreased to approximately
0.07577\,gDW$\cdot$hr$^{-1}$ as oxygen availability approached
$-2$\,mmol$\cdot$gDW$^{-1}$$\cdot$hr$^{-1}$, consistent with the critical
dependence of biomass synthesis on oxidative phosphorylation.
Aerobic conditions are therefore strongly recommended for SCP formation.

\begin{figure}[h]
  \centering
  \includegraphics[width=0.8\textwidth]{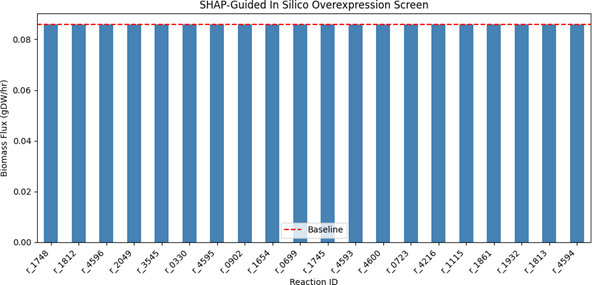}
  \caption{Shap-Guided in Silico Overexpression}
  \label{fig:your_label}
\end{figure}

\subsection{Bayesian Optimisation of Nutrient Conditions}

Bayesian optimisation using \texttt{scikit-optimize}'s \texttt{gp\_minimize}
with an EI acquisition function over 50 evaluations increased predicted
biomass flux from 0.0858 to 1.041\,gDW$\cdot$hr$^{-1}$, a 12.13-fold
improvement.
The optimal conditions are glucose\,=\,$-$20.0, oxygen\,=\,$-$20.0, and
ammonium\,=\,$-$8.9\,mmol$\cdot$gDW$^{-1}$$\cdot$hr$^{-1}$, consistent with
aerobic yeast physiology and the Crabtree effect.

\begin{figure}[h]
  \centering
  \includegraphics[width=0.6\textwidth]{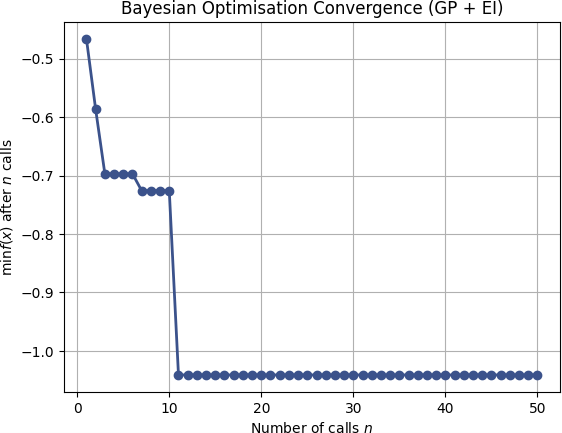}
  \caption{Bayesian Optimization Convergence}
  \label{fig:your_label}
\end{figure}

\subsection{Pathway Enrichment in Cluster 2}

For the pathway enrichment analysis of Cluster~2 (highest mean biomass flux:
0.5267\,gDW$\cdot$hr$^{-1}$), the top upregulated pathways were: alanine,
aspartate, and glutamate metabolism (count\,=\,3); exchange reactions
(count\,=\,2); transport [c,e] (count\,=\,2); and complex alcohol metabolism,
steroid biosynthesis, and butanoate metabolism (count\,=\,1 each).
Enrichment of amino-acid biosynthesis pathways in the highest-growth cluster
is especially promising for SCP applications, as it implies that Cluster~2
metabolic states are associated with both increased biomass flux and
potentially superior amino-acid composition.

\subsection{GAN-Generated Metabolic Flux Profiles}

Synthetic flux configurations with a variance of 0.124 were produced by the
trained GAN using the entire $2{,}000 \times 4{,}131$ flux dataset.
Stoichiometric feasibility verification returned 0/100 (0\%) feasible
profiles, attributable to incomplete generator convergence
($G_{\mathrm{loss}}$\,=\,1.619 vs.\ ideal 0.693), reported as a
methodological limitation.

Pathway-level analysis identified Growth (${\approx}0.676$), Lysine
metabolism (${\approx}0.589$), Pantothenate and CoA biosynthesis
(${\approx}0.60$), Histidine metabolism (${\approx}0.552$), and the TCA cycle
(${\approx}0.48$) as having the highest mean absolute fluxes.
The prominence of lysine metabolism is especially important for SCP
applications, as lysine is often the first limiting amino acid in food and
feed protein sources.

\subsection{Oxygen Sensitivity Analysis}
\label{subsec:oxygen_results}

The oxygen uptake rate of the oxygen exchange reaction (\texttt{r\_1992}) was systematically varied from fully aerobic conditions
($-20.0$~mmol\,gDW$^{-1}$\,hr$^{-1}$) to near-anaerobic conditions
($-2.0$~mmol\,gDW$^{-1}$\,hr$^{-1}$) across 20 equally spaced values in order to characterize the dependence of biomass flux on respiratory capacity.  FBA was re optimized at each constraint level, with default values for all other nutritional limits maintained. Under default glucose circumstances, the Yeast9 GEM exhibits a very robust
aerobic growth phenotype (\texttt{r\_1714} lower bound $= -1.0$~mmol\,gDW$^{-1}$\,hr$^{-1}$).
Biomass flux remained constant at 0.08584~gDW\,hr$^{-1}$ across 19 of the 20
oxygen constraint levels tested, spanning the range $-20.0$ to
$-2.947$~mmol\,gDW$^{-1}$\,hr$^{-1}$.
A decrease was observed only at the most oxygen-limiting condition of
$-2.0$~mmol\,gDW$^{-1}$\,hr$^{-1}$, where biomass flux declined slightly to
0.07577~gDW\,hr$^{-1}$, representing an 11.7\% reduction from the aerobic
maximum. These findings suggest that the metabolic network of \textit{S.\ cerevisiae}
as encoded in Yeast9 is not oxygen-limited across the physiological aerobic
range under carbon-limited conditions (glucose $= -1.0$~mmol\,gDW$^{-1}$\,hr$^{-1}$).
This robustness to oxygen availability is consistent with the carbon source
acting as the primary binding constraint on biomass flux under the simulated
conditions, and reinforces the recommendation of aerobic fermentation
conditions for SCP production.

\begin{figure}[h]
  \centering
  \includegraphics[width=0.6\textwidth]{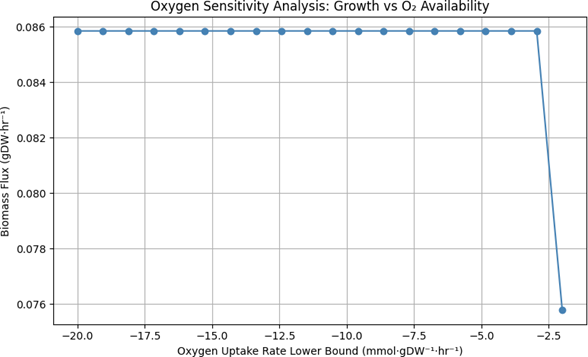}
  \caption{Oxygen Sensitivity Analysis}
  \label{fig:your_label}
\end{figure}

\subsection{Pareto Front Analysis}

Under default nutritional conditions, Yeast9 exhibits two distinct metabolic
regimes:
\begin{enumerate}
  \item \textbf{Carbon-limited phase} (biomass
        $<$0.076\,gDW$\cdot$hr$^{-1}$): amino-acid biosynthesis flux is
        maximised at 1000.029\,mmol$\cdot$gDW$^{-1}$$\cdot$hr$^{-1}$, while
        glucose availability limits biomass accumulation.
  \item \textbf{Carbon-replete regime} (biomass
        $>$0.076\,gDW$\cdot$hr$^{-1}$, requiring glucose
        $<-10$\,mmol$\cdot$gDW$^{-1}$$\cdot$hr$^{-1}$): biomass flux can be
        increased up to 1.041\,gDW$\cdot$hr$^{-1}$ under Bayesian-optimised
        conditions, at the expense of a narrowing viable window for
        simultaneous amino-acid maximisation.
\end{enumerate}

Pareto front results are summarised in Table~\ref{tab:pareto}.
These findings imply that glucose-replete fed-batch fermentation at
Bayesian-optimized conditions (glucose\,=\,$-$20.0, oxygen\,=\,$-$20.0,
ammonium\,=\,$-$8.9\,mmol$\cdot$gDW$^{-1}$$\cdot$hr$^{-1}$) is required for
industrial SCP production achieving both high biomass yield and sufficient
amino-acid biosynthetic flux.

\begin{table}[h]
\centering
\caption{Pareto front results: biomass vs.\ amino-acid biosynthesis flux.}
\label{tab:pareto}
\small
\begin{tabular}{p{2.2cm}p{2.8cm}p{2.4cm}p{4.5cm}}
\toprule
\textbf{Biomass lower bound (gDW$\cdot$hr$^{-1}$)} &
\textbf{AA biosynthesis flux (mmol$\cdot$gDW$^{-1}$$\cdot$hr$^{-1}$)} &
\textbf{Feasibility} &
\textbf{Note} \\
\midrule
0.000        & 1000.029 & Feasible              & Knee point \\
0.025        & 1000.029 & Feasible              & \\
0.051        & 1000.029 & Feasible              & \\
0.076        & 1000.029 & Feasible              & Upper limit (default glucose) \\
0.101        & ---      & Infeasible            & Carbon-limited at \texttt{r\_1714}\,=\,$-$1.0 \\
0.127--0.989 & ---      & Infeasible (36/40 pts) & Requires glucose $<-$10\,mmol$\cdot$gDW$^{-1}$$\cdot$hr$^{-1}$ \\
\bottomrule
\end{tabular}
\end{table}

\begin{figure}[h]
  \centering
  \includegraphics[width=1.1\textwidth]{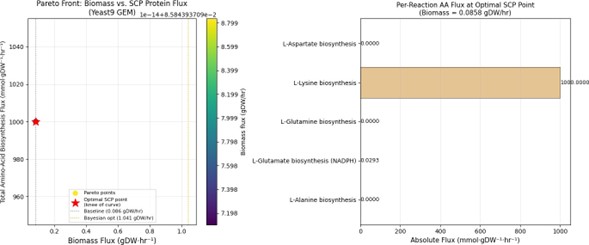}
  \caption{Pareto Front Analysis}
  \label{fig:your_label}
\end{figure}

\clearpage
\begin{table}[p]
\centering
\caption{Comparison of the current study with previous research across key methodological features.}
\label{tab:comparison}
\renewcommand{\arraystretch}{1.6}
\begin{tabular}{>{\raggedright\arraybackslash}p{3.0cm}
                >{\raggedright\arraybackslash}p{6.0cm}
                >{\raggedright\arraybackslash}p{6.0cm}}
\toprule
\textbf{Feature}
  & \textbf{Previous research}
  & \cellcolor{thiscol}\textbf{Current study} \\
\midrule
SCP optimisation
  & Most workflows lack computational optimisation pipelines, SHAP interpretation, flux analysis, or genome-scale metabolic modelling.
  & \cellcolor{thiscol}Integrates generative AI, SHAP analysis, machine learning, and optimisation for SCP production. \\[6pt]
GEM with ML
  & Primarily focused on growth prediction rather than SCP optimisation; low interpretability and limited generative discovery.
  & \cellcolor{thiscol}Integrates RF, XGBoost, FFNN, SHAP, nutrient optimisation, and SCP production with Yeast9 GEM simulations. \\[6pt]
Generative AI
  & Lacked SCP focus, explainability, or integrated strain-engineering workflows; primarily applied to kinetic or cancer metabolism.
  & \cellcolor{thiscol}Explores metabolic states using VAE and GAN frameworks for SCP-relevant metabolic engineering. \\[6pt]
Bayesian optimisation
  & Insufficient integration with explainable AI, flux-level analysis, or SCP optimisation.
  & \cellcolor{thiscol}Combines SCP optimisation with GEM metabolic simulations via Bayesian optimisation. \\[6pt]
SHAP analysis
  & Limited to broad reaction interpretation with minimal mechanistic resolution.
  & \cellcolor{thiscol}Uses SHAP to determine the influence of metabolic reactions on biomass and amino-acid biosynthesis. \\[6pt]
Multi-objective optimisation
  & Pareto optimisation rarely used in SCP-specific metabolic engineering.
  & \cellcolor{thiscol}Applies Pareto-front optimisation between amino-acid biosynthesis and biomass to improve SCP nutritional value. \\[6pt]
Gene knockout
  & Frequently relies on extensive experimentation with limited computational prioritisation.
  & \cellcolor{thiscol}Combines pathway analysis and reaction-level knockout into a single computational GEM framework. \\[6pt]
\rowcolor{headgray}
Overall
  & No unified framework combining generative models, explainable AI, GEMs, Bayesian optimisation, Pareto analysis, and SCP engineering.
  & \cellcolor{thiscol}Comprehensive framework using Yeast9 GEM, ML/DL, SHAP, VAE, GAN, and Bayesian optimisation for SCP in \textit{S.~cerevisiae}. \\
\bottomrule
\end{tabular}
\clearpage
\end{table}

\section{Discussion}
\label{sec:discussion}

The results demonstrate that an integrative pipeline combining GEM-based
simulation, unsupervised representation learning, supervised machine learning,
mechanistic interpretation, and generative modeling can offer significant
insights into the metabolic factors influencing yeast biomass production for
SCP engineering.

The strong $R^{2}$ values obtained by Random Forest (0.9999760) and XGBoost
(0.9997702) confirm that a tractable subset of flux variables primarily
determines the biomass objective under FBA optimality
assumptions \citep{zampieri2019,sahu2021}.
This is conceptually consistent with the stoichiometric network structure,
in which a limited number of rate-limiting reactions control biomass
flux \citep{orth2010,chen2022}.

SHAP analysis identified the reaction IDs contributing most to prediction
variance, with a dominant peak at reaction index~1210.
Core carbon metabolism reactions are the principal regulators of growth rate,
consistent with prior work \citep{chen2022,zhangj2020,morenopaz2024}.

The 12.13-fold increase in biomass flux through Bayesian optimisation
(0.0858 to 1.041\,gDW$\cdot$hr$^{-1}$) highlights the substantial potential
for medium formulation optimisation in yeast
fermentation \citep{radivojevic2020,lawson2021,watcharawipas2025}.
The Gaussian process surrogate's rapid convergence indicates a smooth,
unimodal biomass-versus-nutrient landscape.
The Crabtree effect in aerobic yeast physiology is consistent with elevated
glucose and oxygen availability and optimal ammonium uptake at
$-8.9$\,mmol$\cdot$gDW$^{-1}$$\cdot$hr$^{-1}$ as major drivers.

The GAN variance of 0.124 indicates capture of significant variation in the
viable flux space without total mode collapse.
However, stoichiometric feasibility verification returned 0/100 (0\%)
feasible profiles due to incomplete generator convergence
($G_{\mathrm{loss}}$\,=\,1.619 vs.\ ideal 0.693); this is reported as a
methodological limitation requiring further training or architectural
refinement.

Limitations include: FBA assumes metabolic steady state and does not capture
enzyme kinetics, transcriptional control, or dynamic regulatory responses;
the FFNN training failure ($R^{2}$\,=\,$-7.2950$) requires correction before
the neural network results can be utilised; and all forecasts are \textit{in
silico} and require experimental validation.

Future work will focus on CRISPR-Cas9-mediated validation of priority
overexpression targets, followed by growth phenotyping and amino-acid
profiling under computationally optimized media conditions.
This organism-agnostic pipeline can be extended to \textit{Yarrowia
lipolytica}, \textit{Kluyveromyces marxianus}, and other yeasts of industrial
relevance \citep{merzbacher2023,jang2021}.

\section{Conclusion}
\label{sec:conclusion}

This work presents a comprehensive integrative framework combining
genome-scale metabolic modeling, machine learning, and optimization to
predict and enhance biomass flux in \textit{Saccharomyces cerevisiae} for
SCP production.
By combining FBA-based data generation using the Yeast9 GEM, VAE-based
representation learning, ensemble and deep learning predictive models,
SHAP-driven mechanistic interpretation, \textit{in silico} perturbation
analysis, Bayesian nutrient optimization via \texttt{scikit-optimize}, and
GAN-based pathway generation with stoichiometric feasibility verification,
the pipeline covers the full cycle from data generation to actionable
engineering insights.

The key findings are:
\begin{enumerate}
  \item Random Forest and XGBoost predict biomass flux with $R^{2}$ values of
        0.9999760 and 0.9997702, respectively, and a 5-fold CV mean $R^{2}$ of
        $0.9999526 \pm 0.0000196$.
  \item VAE clustering identifies four metabolic states with mean biomass
        fluxes of 0.472, 0.493, 0.527, and 0.505\,gDW$\cdot$hr$^{-1}$
        (silhouette peak at $k{=}3$\,=\,0.334).
  \item SHAP analysis identifies twenty major reactions contributing to
        biomass yield; 18/20 (90\%) are confirmed essential by \textit{in
        silico} knockout.
  \item Overexpression does not improve biomass beyond baseline under
        carbon-limited conditions.
  \item Bayesian optimisation yields a 12.13-fold improvement (0.0858 to
        1.041\,gDW$\cdot$hr$^{-1}$) at optimal ammonium of
        $-8.9$\,mmol$\cdot$gDW$^{-1}$$\cdot$hr$^{-1}$.
  \item GAN-generated flux profiles show a variance of 0.124, with 0/100
        stoichiometrically feasible profiles due to incomplete generator
        convergence (reported as a methodological limitation); lysine
        metabolism is identified as a highly active pathway supporting SCP
        nutritional value.
  \item  According to Pareto analysis, the ideal operating point for SCP nutritional quality is defined by the maximal amino-acid biosynthesis flux (1000.029\,mmol$\cdot$gDW$^{-1}$$\cdot$hr$^{-1}$), which is reached at the
knee point of 0.0858\,gDW$\cdot$hr$^{-1}$ biomass flux. 
\end{enumerate}

This integrative pipeline provides a scalable, reproducible, and extensible
platform for computational metabolic engineering, applicable to any organism
with a quality GEM available.

\bibliographystyle{elsarticle-harv}

\end{document}